\documentclass[11pt,a4paper]{article}
\usepackage[hyperref]{acl2020}
\usepackage{times}
\usepackage{latexsym}
\usepackage{graphicx}

\usepackage{microtype}

\aclfinalcopy % Uncomment this line for the final submission
\title{Generating Followup Questions for Interpretable 
       Multi-hop Question Answering}

\author{Christopher Malon \\
  NEC Laboratories America \\
  Princeton, NJ 08540 \\
  \texttt{malon@nec-labs.com} \\\And
  Bing Bai \\
  NEC Laboratories America \\
  Princeton, NJ 08540 \\
  \texttt{bbai@nec-labs.com} \\}

\date{}

\begin{document}
\maketitle
\begin{abstract}
We propose a framework for answering open domain multi-hop questions
in which partial information is read and used to generate followup questions,
to finally be answered by a pretrained single-hop answer extractor.
This framework makes each hop interpretable, and makes the retrieval associated
with later hops as flexible and specific as for the first hop.
As a first instantiation of this framework, we train a pointer-generator
network to predict followup questions based on the question and partial
information.  This provides a novel application of a neural question
generation network, which is applied to give weak ground truth single-hop
followup questions based on the final answers and their supporting facts.
Learning to generate followup questions that select the relevant answer spans
against downstream supporting facts, while avoiding distracting premises, poses
an exciting semantic challenge for text generation.  We present an evaluation
using the two-hop bridge questions of HotpotQA.
\end{abstract}

\section{Introduction}

% Although early work on multi-hop question answering in Wikihop and Medhop
% focused on relation extraction \cite{wikihop} using a document corpus,
% HotpotQA \cite{hotpotqa} introduced a dataset where both the questions
% and the supporting documents are unstructured
% text, and supporting facts are labeled.

Multi-hop question answering tests the ability of a system to retrieve and
combine multiple facts to answer a single question.
HotpotQA \cite{hotpotqa} introduces a task where questions are free-form
text, supporting facts come from Wikipedia, and answer text and
supporting facts are labeled.  The questions in HotpotQA
are further categorized as bridge-type questions
or comparison-type questions.  For comparison questions, often
all necessary facts may be retrieved using terms in the question itself.
For challenging
bridge-type questions, it may not be possible to retrieve all the necessary
facts based on the terms present in the original question alone.
Rather, partial information must first be retrieved and used to
formulate an additional query.

Although many systems have been submitted to the HotpotQA leaderboard,
surprisingly, only a few have directly addressed the challenge of
followups.  Systems can either be evaluated in a distractor setting,
where a set of ten paragraphs containing all supporting facts is provided,
or in a full wiki setting, where supporting facts must be retrieved from
all of Wikipedia.  The systems that compete only in the distractor
setting % list them here?
can achieve good performance by combining and ranking the information
provided, without performing followup search queries.
Furthermore, even in the distractor setting, \citet{singlehop} found that
only 27\% of the questions required multi-hop reasoning, because additional
evidence was redundant or unnecessary or the distractors were weak.
They trained a single-hop model that considered each paragraph in isolation
and ranked confidences of the answers extracted from each, to obtain
competitive performance.

Of the nine systems with documentation submitted to the full wiki HotpotQA
leaderboard as of 24 November 2019, 
four of them \citep{nie, ye, nishida, hotpotqa}
attempt to retrieve all relevant data with one search based on the
original question, without any followups.
\citet{fang} retrieves second hop paragraphs simply by following hyperlinks
from or to the first hop paragraphs.

\citet{iterativequery}, \citet{cognitivegraph}, and \citet{muppet}
form various kinds of followup queries without writing a new question
to be answered.  \citet{iterativequery} trains a span extractor to predict
the longest common subsequence between the question plus the first hop evidence
and the (unseen) second hop evidence.  At inference time, these predicted
spans become followup search queries.  In \citet{cognitivegraph}, 
a span extractor is trained using the titles of the second hop evidence.
\citet{muppet} trains a neural retrieval model that uses maximum inner
product with an encoding of the question plus first hop evidence to retrieve
second hop evidence.

% We haven't mentioned which methods reevaluate all info with respect
% to the original question right before the end.

\citet{qdecomp} forms not just followup queries but followup questions.
They use additional specially labeled data to train a pointer network to
divide the original question into substrings, and use handcrafted rules
to convert these substrings into subquestions.  The original question
is answered by the second subquestion, which incorporates a substitution
of the answer to the first subquestion.
% Closest to our approach, \citet{qdecomp} decomposes an original question
% into subquestions, and answers a followup question formed using the answer
% to the first question.  The subquestion formation process
% is trained by additional specially labeled data indicating substrings
% of the original question, and then uses handcrafted rules to make questions
% from these substrings.

While performing followup retrievals of some sort should be essential for
correctly solving the most difficult multi-hop problems, formulating a
followup {\em question} whose answer becomes the answer to the original
question is motivated primarily by interpretability rather than accuracy.
% Even so, the flexibility of forming questions that are not limited to
% spans of the 
In this paper, we pursue a trained approach to generating followup questions
that is not bound by handcrafted rules, posing a new and challenging
application for abstractive summarization and neural question generation
technologies.  Our contributions are to define the task of a followup
generator module (Section 2), to propose a fully trained solution to
followup generation (Section 3), and to establish an objective evaluation
of followup generators (Section 5).

\section{Problem Setting}

Our technique is specifically designed to address the challenge of 
discovering new information is needed that is not specified by the terms of
the original question.  At the highest level, comparison questions do
not pose this challenge, because each quantity to be compared is specified
by part of the original question.  (They also pose different semantics
than bridge questions because a comparison must be applied after retrieving
answers to the subquestions.)  Therefore we focus only on bridge questions
in this paper.

\begin{figure}[t]
\centering
\includegraphics[width=0.44\textwidth]{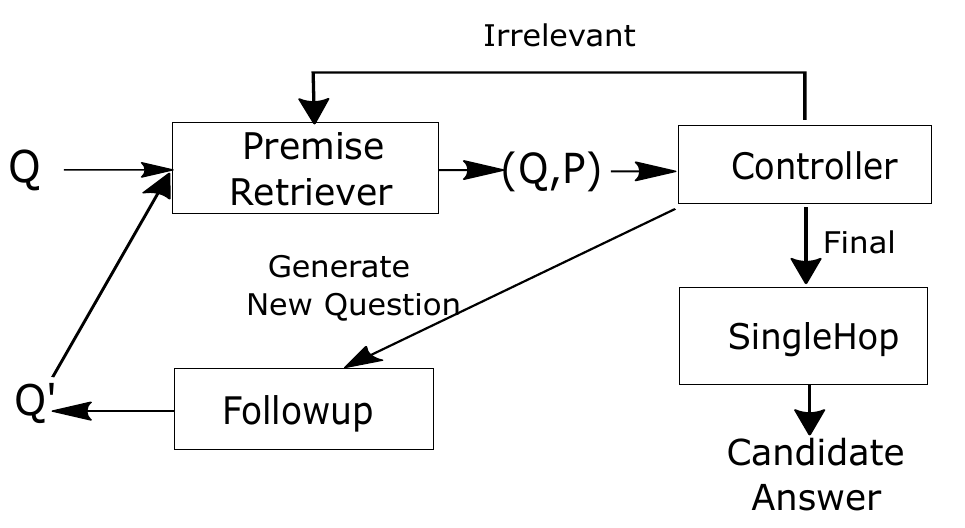}
\caption{The architecture of our system to generate intermediate questions
for answer extraction. \label{fig:structure}
}
\end{figure}

Figure~\ref{fig:structure} shows our pipeline to answer a multi-hop
bridge question.
As partial information is obtained, an original question is iteratively
reduced to simpler questions generated at each hop.  Given an input question
or subquestion, possible premises which may answer the subquestion are
obtained from an information retrieval module.  Each possible premise
is classified against the question as irrelevant, containing a final answer,
or containing intermediate information, by a three-way controller module.
% To enable training on two hops but application to arbitrarily many hops,
% the controller uses only the most recently generated question with the
% premise.
For premises that contain a final answer, the answer is extracted with
a single-hop question answering module.  For premises that contain intermediate
information, a question generator produces a followup question, and the
process may be repeated with respect to this new question.
It is this question generator that is the focus of this paper.
Various strategies may be used to manage the multiple reasoning paths
that may be produced by the controller.  Details are in section~\ref{sec:exper}.

Although our method applies to bridge questions with arbitrary number of hops,
for simplicity we focus on two-hop problems and on training the
followup question generator.
Let $Cont$ denote the controller,
$SingleHop$ denote the answer extractor, and $Followup$ denote
the followup generator.
Let $Q_1$ be
a question with answer $A$ and gold supporting premises
$\hat{P_1}$ and $\hat{P_2}$, and suppose that $\hat{P_2}$ but not $\hat{P_1}$
contains the answer.
% Assuming the controller and answer extractor function perfectly,
The task of the followup generator is to use $Q_1$ and $\hat{P_1}$ to
generate a followup question $Q_2$ such that
\begin{eqnarray}
SingleHop(Q_2, \hat{P_2}) & = & A \label{eq:rightanswer} \\
Cont(Q_2, \hat{P_2}) & = & Final \label{eq:recognized} \\
\mbox{and}\; Cont(Q_2, P) & = & Irrel \, \mbox{for}\, P \neq \hat{P_2} \ldotp \label{eq:irrelevant}
\end{eqnarray}
Failure of any of these desiderata could harm label accuracy in the
HotpotQA full wiki or distractor evaluations.

Some questions labeled as bridge type in HotpotQA have a different logical
structure, called ``intersection'' by \citet{qdecomp}.  Here the subquestions
specify different properties that the answer entity is supposed to satisfy,
and the intersection of possible answers to the subquestions is the answer
to the original question.  Our approach is not oriented towards this type
of question, but there is no trivial way to exclude them from the dataset.

One non-interpretable implementation of our pipeline would be
for $Followup$ to simply output $Q_1$ concatenated with $P_1$
as the ``followup question.''  Then $SingleHop$ would operate on input
that really does not take the form of a single question, along with $P_2$,
to determine the final answer.  Effectively, $SingleHop$ would be doing
multi-hop reasoning.  To ensure that $Followup$ gets credit only
for forming real followup questions, we insist that
$SingleHop$ is first trained as a single-hop answer extractor,
by training it on SQuAD 2.0 \citep{squad}, then freeze it while $Followup$ and
$Cont$ are trained.

\section{Method}

\label{sec:method}

Ideally, we might train $Followup$ using cross entropy losses inspired
by equations \ref{eq:rightanswer}, \ref{eq:recognized},
and \ref{eq:irrelevant} with $SingleHop$ and $Cont$ fixed, but
the decoded output $Q_2$ is not differentiable with respect to $Followup$
parameters.  Instead, we train $Followup$ with a token-based loss against
a set of weakly labeled ground truth followup questions.

The weakly labeled ground truth followups are obtained using a neural
question generation (QG) network.  Given a context $\overline{C}$ and an
answer $\overline{A}$,
QG is the task of finding a question
\begin{equation}
\overline{Q} = \mbox{argmax}_Q Prob(Q | \overline{C}, \overline{A})
\end{equation}
most likely to have produced it.  We use reverse SQuAD to train the QG model
of \citet{s2sa}, which performs near the top of an extensive set of models
tested by \citet{qgeneration} and has an independent implementation available.
Applied to our training set
with $\overline{C} = \hat{P_2}$ and $\overline{A} = A$, it gives
us a weak ground truth followup $\overline{Q_2}$.

We instantiate the followup question generator, which uses $Q_1$ and $P_1$
to predict $Q_2$, with a pointer-generator
network \citep{see}.  This is a sequence to sequence model whose decoder
repeatedly chooses between generating a word from a fixed vocabulary and
copying a word from the input.  Typically, pointer-generator networks are used
for abstractive summarization.
Although the output serves a different role here, their copy
mechanism is useful in constructing a followup that uses information from
the original question and premise.

We train $Cont$ with cross-entropy loss
for ternary classification on the ground truth triples
$(Q_1, \hat{P_1}, Intermediate)$,
$(Q_1, \hat{P_2}, Final)$ if $SingleHop(Q_1, \hat{P_2}) \cap A \neq \emptyset$,
and $(Q_1, P, Irrel)$ for all other $P$.
In this way the controller learns to predict when a premise has sufficient
or necessary information to answer a question.
Both $Cont$ and
$SingleHop$ are implemented by BERT following the code by \citet{devlin}.

\section{Related Work}

Evaluating a followup question generator by whether its questions are
answered correctly is analogous to verifying the factual accuracy of
abstractive summarizations, which has been studied by many, including
\citet{falke}, who estimate factual correctness using a natural
language inference model, and find that it does not correlate
with ROUGE score.  Contemporaneous work by \citet{radiology}
uses feedback from a fact extractor in reinforcement learning to
optimize the correctness of a summary, suggesting an interesting
future direction for our work.

A recent neural question generation model has incorporated
feedback from an answer extractor into the training of a question generator,
rewarding the generator for constructing questions the extractor can
answer correctly \citep{sap}.  Although the loss is not backpropagated
through both the generator and extractor, the generator is penalized by
token level loss against ground truth questions when the question is answered
wrongly, but by zero loss when it constructs a variant that the extractor
answers correctly.

\section{Experiments}

\label{sec:exper}

To isolate the effect of our followup generator on the types of questions
for which it was intended, our experiments cover the subset of questions in HotpotQA labeled with exactly two supporting facts, with the answer string
occurring in exactly one of them.  There are 38,806 such questions for
training and 3,214 for development, which we use for testing because
the structure of the official test set is not available.
For a baseline we compare to a trivial followup generator
that returns the original question $Q_1$ without any rewriting.

\begin{table}[htb]
\centering
\begin{tabular}{lcc}
\hline
& EM & F1 \\
\hline
{\em Oracle setting} & & \\
Trained $Q_2$ & 14.7 & 19.0 \\
$Q_2 = Q_1$ & 27.6 & 34.9 \\
$Q_1$ else $Q_2$ & {\bf 34.7} & {\bf 43.8} \\
\hline
{\em Full system} & & \\
One hop ($Q_1$ only) & 16.8 & 21.5 \\
Two hops (trained $Q_2$) & {\bf 19.8} & {\bf 25.4} \\
\hline
\end{tabular}
\caption{Answer accuracy on filtered subset of HotpotQA development set in the
distractor setting.}
\label{tab:eval}
\end{table}

\begin{table*}[t]
\centering
\begin{tabular}{p{1.5in}p{.75in}p{.75in}p{1.5in}p{.75in}}
\hline
$Q_1$ & Truth & $Q_1$ Answer & $Q_2$ & $Q_2$ Answer \\
\hline
Randall Cunningham II was a multi-sport athlete at the high school
located in what Nevada city? & Summerlin & --- &
where is bishop gorman high school located? & Summerlin, Nevada \\
Alexander Kerensky was defeated and destroyed by the Bolsheviks in
the course of a civil war that ended when? & October 1922 & --- &
what was the name of the russian civil war? & The Russian Civil War \\
Peter Marc Jacobson is best known as the co-creator of the popular
sitcom "The Nanny", which he created and wrote with his then wife an actress
born in which year ? & 1957 & 1993 &
what year was fran drescher born in? & 1957 \\
Who did the Star and Dagger bass player marry? & Sean Yseult. & Sean Yseult &
what was the name of the american rock musician? & Chris Lee \\
\hline
\end{tabular}
\caption{Example generated followup questions $Q_2$,
evaluated against oracle $\hat{P_2}$.}
\label{tab:samples}
\end{table*}

First, we evaluate performance using an oracle controller,
which forwards only $(Q_1, \hat{P_1})$ to the followup generator,
and only $(Q_2, \hat{P_2})$ to the answer extractor.
Results are shown in Table~\ref{tab:eval}.
Best performance is achieved using the system ``$Q_1$ else $Q_2$,''
which answers with $SingleHop(Q_1, \hat{P_2})$ or $SingleHop(Q_2, \hat{P_2})$,
whichever is non-null.
Thus, although many questions are really single-hop and best answered using
the original question, using the followup questions when a single-hop answer
cannot be found helps the F1 score by 8.9\%.
Table~\ref{tab:samples} shows followup generations and extracted answers
in two typical successful and two typical failed cases.

Next we consider the full system of Figure~\ref{fig:structure}.
We use the distractor paragraphs provided.
We run the loop for up to two hops, collecting all
answer extractions requested by the
controller, stopping after the first hop where a non-null extracted answer was
obtained.  If multiple extractions were requested for the same problem,
we take the answer in where $SingleHop$ had the highest confidence.
The controller requested 2,989 followups, and sent 975 $(Q, P)$ pairs
for answer extraction in hop one, and 1,180 in hop two.  The performance
gain shows that the followup generator often can generate questions
which are good enough for the frozen single hop model to understand and
extract the answer with, even when the question must be specific enough
to avoid distracting premises.

\section{Conclusion}

Followup queries are essential to solving the difficult cases of
multi-hop QA, and real followup questions are an advance
in making this process interpretable.  We have shown that
pointer generator networks can effectively learn to read partial information
and produce a fluent, relevant question about what is not known,
which is a complement to their typical role in summarizing what is known.
Our task poses a novel challenge that tests semantic properties of
the generated output.

By using a neural question generator to produce weak ground truth followups,
we have made this task more tractable.  Future work should examine using
feedback from the answer extractor or controller to improve the
sensitivity and specificity of the generated followups.
Additionally, the approach should be developed on new datasets such as
QASC \citep{qasc}, which are designed to make single-hop retrieval less
effective.

% We answer a question of Gatt&Krahmer, Novikova
% (both cited in section 3.2 of Min "Question Decomposition"
% who say this is a difficult task

\bibliography{followup}

\begin{thebibliography}{19}
\expandafter\ifx\csname natexlab\endcsname\relax\def\natexlab#1{#1}\fi

\bibitem[{Devlin et~al.(2019)Devlin, Chang, Lee, and Toutanova}]{devlin}
Jacob Devlin, Ming-Wei Chang, Kenton Lee, and Kristina Toutanova. 2019.
\newblock \href {https://doi.org/10.18653/v1/N19-1423} {{BERT}: Pre-training of
  deep bidirectional transformers for language understanding}.
\newblock In \emph{Proceedings of the 2019 Conference of the North {A}merican
  Chapter of the Association for Computational Linguistics: Human Language
  Technologies, Volume 1 (Long and Short Papers)}, pages 4171--4186,
  Minneapolis, Minnesota. Association for Computational Linguistics.

\bibitem[{Ding et~al.(2019)Ding, Zhou, Chen, Yang, and Tang}]{cognitivegraph}
Ming Ding, Chang Zhou, Qibin Chen, Hongxia Yang, and Jie Tang. 2019.
\newblock \href {https://doi.org/10.18653/v1/P19-1259} {Cognitive graph for
  multi-hop reading comprehension at scale}.
\newblock In \emph{Proceedings of the 57th Annual Meeting of the Association
  for Computational Linguistics}, pages 2694--2703, Florence, Italy.
  Association for Computational Linguistics.

\bibitem[{Falke et~al.(2019)Falke, Ribeiro, Utama, Dagan, and Gurevych}]{falke}
Tobias Falke, Leonardo F.~R. Ribeiro, Prasetya~Ajie Utama, Ido Dagan, and Iryna
  Gurevych. 2019.
\newblock \href {https://doi.org/10.18653/v1/P19-1213} {Ranking generated
  summaries by correctness: An interesting but challenging application for
  natural language inference}.
\newblock In \emph{Proceedings of the 57th Annual Meeting of the Association
  for Computational Linguistics}, pages 2214--2220, Florence, Italy.
  Association for Computational Linguistics.

\bibitem[{Fang et~al.(2019)Fang, Sun, Gan, Pillai, Wang, and Liu}]{fang}
Yuwei Fang, Siqi Sun, Zhe Gan, Rohit Pillai, Shuohang Wang, and Jingjing Liu.
  2019.
\newblock Hierarchical graph network for multi-hop question answering.
\newblock \emph{CoRR}, 1911.03631.

\bibitem[{Feldman and El-Yaniv(2019)}]{muppet}
Yair Feldman and Ran El-Yaniv. 2019.
\newblock \href {https://doi.org/10.18653/v1/P19-1222} {Multi-hop paragraph
  retrieval for open-domain question answering}.
\newblock In \emph{Proceedings of the 57th Annual Meeting of the Association
  for Computational Linguistics}, pages 2296--2309, Florence, Italy.
  Association for Computational Linguistics.

\bibitem[{Khot et~al.(2019)Khot, Clark, Guerquin, Jansen, and Sabharwal}]{qasc}
Tushar Khot, Peter Clark, Michal Guerquin, Peter Jansen, and Ashish Sabharwal.
  2019.
\newblock Qasc: A dataset for question answering via sentence composition.
\newblock \emph{CoRR}, 1910.11473.

\bibitem[{Klein and Nabi(2019)}]{sap}
Tassilo Klein and Moin Nabi. 2019.
\newblock Learning to answer by learning to ask: Getting the best of gpt-2 and
  bert worlds.
\newblock \emph{CoRR}, 1911.02365.

\bibitem[{Min et~al.(2019{\natexlab{a}})Min, Wallace, Singh, Gardner,
  Hajishirzi, and Zettlemoyer}]{singlehop}
Sewon Min, Eric Wallace, Sameer Singh, Matt Gardner, Hannaneh Hajishirzi, and
  Luke Zettlemoyer. 2019{\natexlab{a}}.
\newblock \href {https://doi.org/10.18653/v1/P19-1416} {Compositional questions
  do not necessitate multi-hop reasoning}.
\newblock In \emph{Proceedings of the 57th Annual Meeting of the Association
  for Computational Linguistics}, pages 4249--4257, Florence, Italy.
  Association for Computational Linguistics.

\bibitem[{Min et~al.(2019{\natexlab{b}})Min, Zhong, Zettlemoyer, and
  Hajishirzi}]{qdecomp}
Sewon Min, Victor Zhong, Luke Zettlemoyer, and Hannaneh Hajishirzi.
  2019{\natexlab{b}}.
\newblock \href {https://doi.org/10.18653/v1/P19-1613} {Multi-hop reading
  comprehension through question decomposition and rescoring}.
\newblock In \emph{Proceedings of the 57th Annual Meeting of the Association
  for Computational Linguistics}, pages 6097--6109, Florence, Italy.
  Association for Computational Linguistics.

\bibitem[{Nie et~al.(2019)Nie, Wang, and Bansal}]{nie}
Yixin Nie, Songhe Wang, and Mohit Bansal. 2019.
\newblock \href {https://doi.org/10.18653/v1/D19-1258} {Revealing the
  importance of semantic retrieval for machine reading at scale}.
\newblock In \emph{Proceedings of the 2019 Conference on Empirical Methods in
  Natural Language Processing and the 9th International Joint Conference on
  Natural Language Processing (EMNLP-IJCNLP)}, pages 2553--2566, Hong Kong,
  China. Association for Computational Linguistics.

\bibitem[{Nishida et~al.(2019)Nishida, Nishida, Nagata, Otsuka, Saito, Asano,
  and Tomita}]{nishida}
Kosuke Nishida, Kyosuke Nishida, Masaaki Nagata, Atsushi Otsuka, Itsumi Saito,
  Hisako Asano, and Junji Tomita. 2019.
\newblock \href {https://doi.org/10.18653/v1/P19-1225} {Answering while
  summarizing: Multi-task learning for multi-hop {QA} with evidence
  extraction}.
\newblock In \emph{Proceedings of the 57th Annual Meeting of the Association
  for Computational Linguistics}, pages 2335--2345, Florence, Italy.
  Association for Computational Linguistics.

\bibitem[{Qi et~al.(2019)Qi, Lin, Mehr, Wang, and Manning}]{iterativequery}
Peng Qi, Xiaowen Lin, Leo Mehr, Zijian Wang, and Christopher~D. Manning. 2019.
\newblock \href {https://doi.org/10.18653/v1/D19-1261} {Answering complex
  open-domain questions through iterative query generation}.
\newblock In \emph{Proceedings of the 2019 Conference on Empirical Methods in
  Natural Language Processing and the 9th International Joint Conference on
  Natural Language Processing (EMNLP-IJCNLP)}, pages 2590--2602, Hong Kong,
  China. Association for Computational Linguistics.

\bibitem[{Rajpurkar et~al.(2018)Rajpurkar, Jia, and Liang}]{squad}
Pranav Rajpurkar, Robin Jia, and Percy Liang. 2018.
\newblock \href {https://doi.org/10.18653/v1/P18-2124} {Know what you don{'}t
  know: Unanswerable questions for {SQ}u{AD}}.
\newblock In \emph{Proceedings of the 56th Annual Meeting of the Association
  for Computational Linguistics (Volume 2: Short Papers)}, pages 784--789,
  Melbourne, Australia. Association for Computational Linguistics.

\bibitem[{See et~al.(2017)See, Liu, and Manning}]{see}
Abigail See, Peter~J. Liu, and Christopher~D. Manning. 2017.
\newblock \href {https://doi.org/10.18653/v1/P17-1099} {Get to the point:
  Summarization with pointer-generator networks}.
\newblock In \emph{Proceedings of the 55th Annual Meeting of the Association
  for Computational Linguistics (Volume 1: Long Papers)}, pages 1073--1083,
  Vancouver, Canada. Association for Computational Linguistics.

\bibitem[{Tuan et~al.(2019)Tuan, Shah, and Barzilay}]{qgeneration}
Luu~Anh Tuan, Darsh~J Shah, and Regina Barzilay. 2019.
\newblock Capturing greater context for question generation.
\newblock \emph{CoRR}, 1910.10274.

\bibitem[{Yang et~al.(2018)Yang, Qi, Zhang, Bengio, Cohen, Salakhutdinov, and
  Manning}]{hotpotqa}
Zhilin Yang, Peng Qi, Saizheng Zhang, Yoshua Bengio, William Cohen, Ruslan
  Salakhutdinov, and Christopher~D. Manning. 2018.
\newblock \href {https://doi.org/10.18653/v1/D18-1259} {{H}otpot{QA}: A dataset
  for diverse, explainable multi-hop question answering}.
\newblock In \emph{Proceedings of the 2018 Conference on Empirical Methods in
  Natural Language Processing}, pages 2369--2380, Brussels, Belgium.
  Association for Computational Linguistics.

\bibitem[{Ye et~al.(2019)Ye, Lin, Liu, Liu, and Sun}]{ye}
Deming Ye, Yankai Lin, Zhenghao Liu, Zhiyuan Liu, and Maosong Sun. 2019.
\newblock Multi-paragraph reasoning with knowledge-enhanced graph neural
  network.
\newblock \emph{CoRR}, 1911.02170.

\bibitem[{Zhang et~al.(2019)Zhang, Merck, Tsai, Manning, and
  Langlotz}]{radiology}
Yuhao Zhang, Derek Merck, Emily~Bao Tsai, Christopher~D. Manning, and Curtis~P.
  Langlotz. 2019.
\newblock Optimizing the factual correctness of a summary: A study of
  summarizing radiology reports.
\newblock \emph{CoRR}, 1911.02541.

\bibitem[{Zhao et~al.(2018)Zhao, Ni, Ding, and Ke}]{s2sa}
Yao Zhao, Xiaochuan Ni, Yuanyuan Ding, and Qifa Ke. 2018.
\newblock \href {https://doi.org/10.18653/v1/D18-1424} {Paragraph-level neural
  question generation with maxout pointer and gated self-attention networks}.
\newblock In \emph{Proceedings of the 2018 Conference on Empirical Methods in
  Natural Language Processing}, pages 3901--3910, Brussels, Belgium.
  Association for Computational Linguistics.

\end{thebibliography}
\bibliographystyle{acl_natbib}

\end{document}